\pdfoutput=1

\documentclass[11pt]{article}

\usepackage[]{acl}

\usepackage{times}
\usepackage{latexsym}

\usepackage{graphicx}

\usepackage[T1]{fontenc}

\usepackage[utf8]{inputenc}
\usepackage{array}

\usepackage{microtype}
\usepackage{multirow}
\usepackage{multicol}
\usepackage{booktabs}

\usepackage{inconsolata}

\usepackage{color}
\usepackage{xspace}
\usepackage{makecell}

\newcommand{\ours}{\textsc{HProPro\ }\xspace}

\title{Exploring Hybrid Question Answering via Program-based Prompting}

\author{Qi Shi\thanks{\ \ Equal Contributions.},\ \ Han Cui$^*$,\ \ Haofeng Wang,\ \ Qingfu Zhu,\ \ Wanxiang Che,\ \ Ting Liu \\
	\normalsize{Research Center for Social Computing and Information Retrieval}\\[-.05cm]
	\normalsize{Harbin Institute of Technology, Harbin, China}\\[-.05cm]
	{\small\tt \{qshi,hcui,hfwang,qfzhu,car,tliu\}@ir.hit.edu.cn}  \\
}

\begin{document}
\maketitle
\begin{abstract}

Question answering over heterogeneous data requires reasoning over diverse sources of data, which is challenging due to the large scale of information and organic coupling of heterogeneous data. 
Various approaches have been proposed to address these challenges. One approach involves training specialized retrievers to select relevant information, thereby reducing the input length. Another approach is to transform diverse modalities of data into a single modality, simplifying the task difficulty and enabling more straightforward processing.
In this paper, we propose \ours, a novel program-based prompting framework for the hybrid question answering task.
\ours follows the code generation and execution paradigm. 
In addition, \ours integrates various functions to tackle the hybrid reasoning scenario.
Specifically, \ours contains function declaration and function implementation to perform hybrid information-seeking over data from various sources and modalities, which enables reasoning over such data without training specialized retrievers or performing modal transformations.
Experimental results on two typical hybrid question answering benchmarks HybridQA and MultiModalQA demonstrate the effectiveness of \ours: it surpasses all baseline systems and achieves the best performances in the few-shot settings on both datasets.

\end{abstract}

\section{Introduction}
\label{sec:intro}

\begin{figure}[t]
    \centering
    \includegraphics[width=0.85\linewidth]{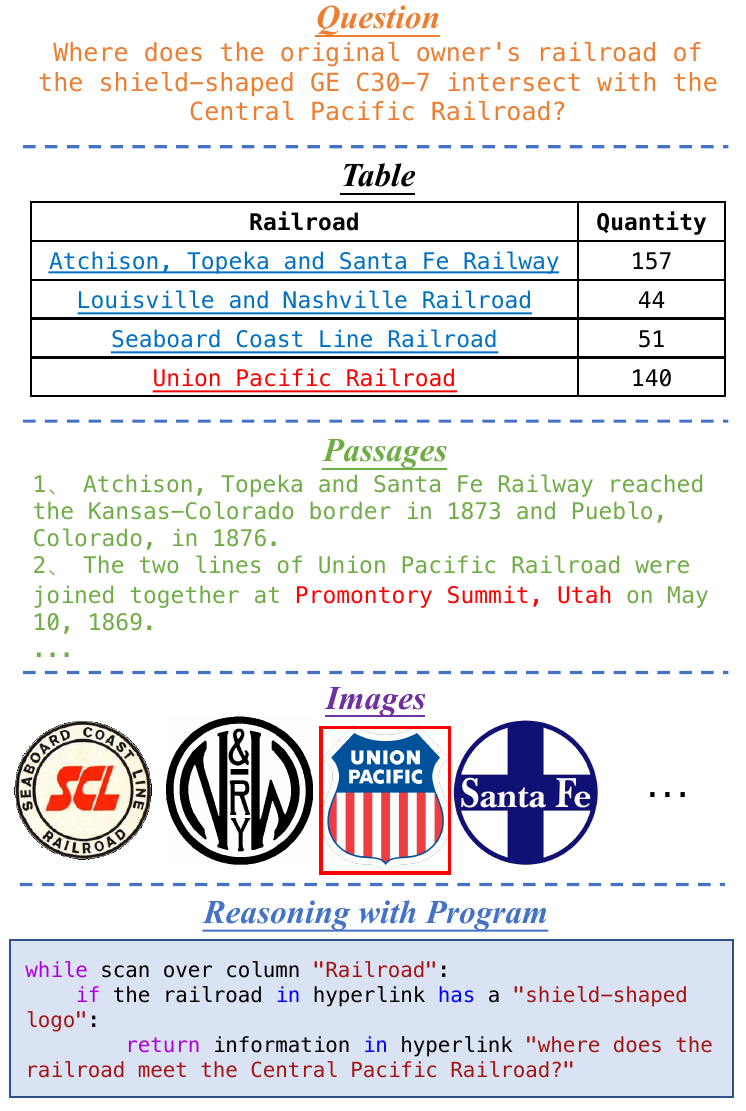}
    \caption{Example of hybrid question answering task with the corresponding program.}
    \label{fig:example}
\end{figure}

Question answering systems \cite{pasupat2015compositional,rajpurkar2016squad,goyal2017making} have attracted significant attention and made considerable progress in recent years. However, real-world data often exists in diverse formats and originates from multiple sources. Consequently, researchers turn their focus to the hybrid question answering (HQA) task \cite{chen2020hybridqa,talmor2020multimodalqa}, which necessitates mixed reasoning across various types of data.
The HQA task is challenging due to the vast amount of information and the organic coupling of heterogeneous data sources. Reasoning over such diverse data requires the ability to understand multiple data types simultaneously. 
For instance, as depicted in Figure \ref{fig:example}, the model must engage in reasoning over both the table and the extensive passages and images linked in hyperlinks to make accurate predictions.

To tackle these challenges, recent approaches focus on training domain-specific models to retrieve or rank elements such as table rows, passages, or images, selecting the most relevant ones to enhance the subsequent reasoning process \cite{eisenschlos2021mate,kumar2021multi,lei2023s}.
Since real-world heterogeneous data is vast and constantly updated, even if these approaches demonstrate promising performance on their focused datasets, their applicability to such intricate data is still limited.
Furthermore, some existing approaches tend to transform diverse modalities of data into a single modality, such as image captioning \cite{cheng2022binding,liu2023mmhqa}, or table-to-text generation \cite{li2021tsqa}, to reduce the task difficulty. 
However, such approaches are constrained by the performance of modal transformation models, which often result in the loss of information.
In a word, these approaches highly rely on data distribution, and the complexity of real-world heterogeneous data makes them exorbitant. 

In contrast to previous approaches, 
we argue that the solution of solving the HQA task should be agnostic to data distribution.
Consequently, we advocate for an optimal solution devising a procedure for determining how to find an answer, rather than merely generating the answer itself.
Noticing that the program could elucidate the reasoning process employed to arrive at the answer (as depicted in Figure \ref{fig:example}), 
in the current era of large language models (LLMs), leveraging a program can serve as an advantageous solution since LLMs are an excellent program generator.
Moreover, the process of program generation necessitates the incorporation of various functions into the program, enabling information-seeking across diverse sources and modalities of data.

\begin{figure*}[htbp]
    \centering
    \includegraphics[width=0.95\linewidth]{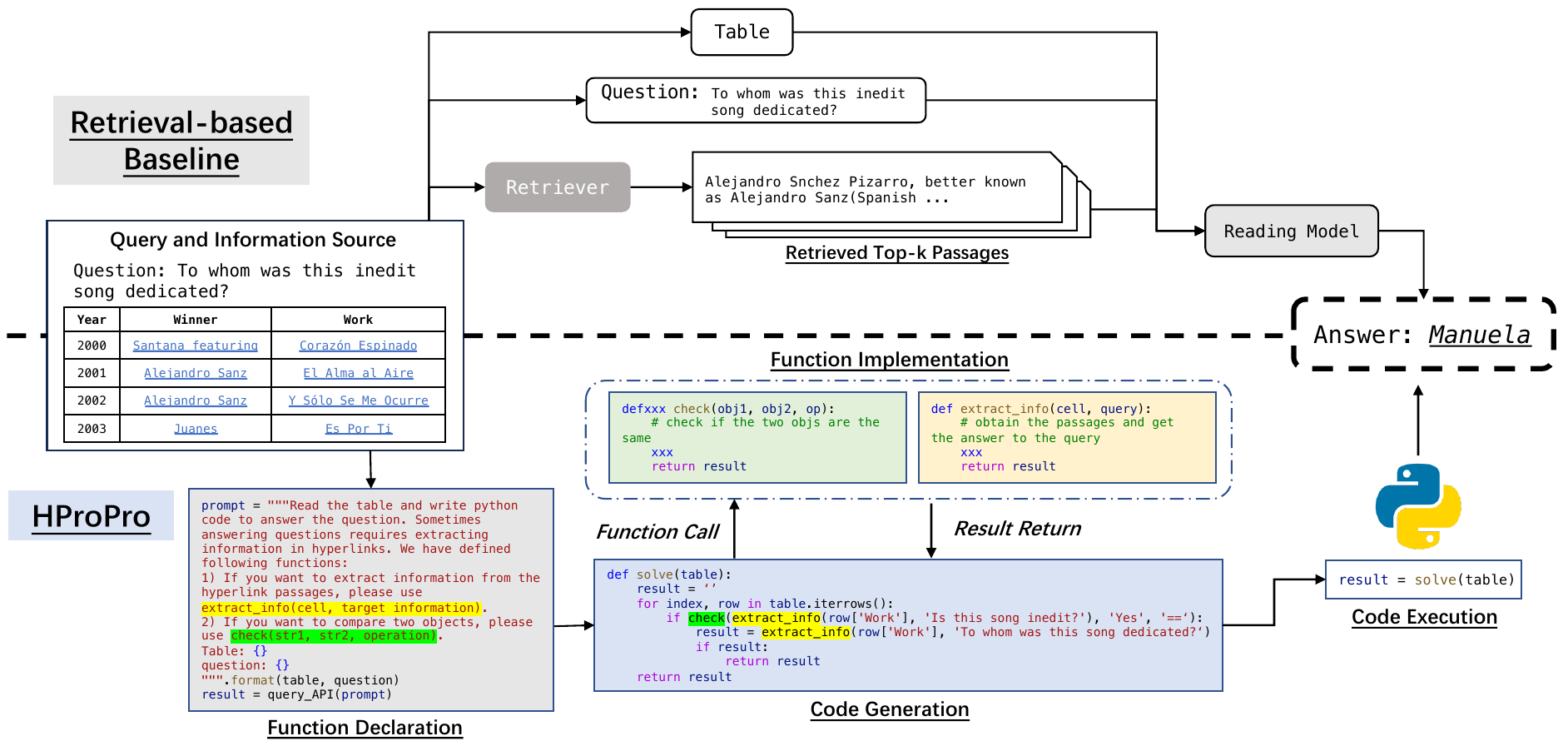}
    \caption{Comparison of \ours with previous retrieval-based methods.}
    \label{fig:framework}
\end{figure*}

Based on the aforementioned considerations, in this paper, we introduce a novel program-based prompting framework \ours (\textbf{H}ybrid \textbf{Pro}gram-Based \textbf{Pro}mpting) for HQA task.
\ours considers the solution as a process of code generation and execution, integrating external customized functions under the few-shot setting\footnote{In this work, we use Python code as the carrier of the program.}.
To facilitate the utilization of customized functions, \ours incorporates two key components: \textbf{Function Declaration} during the code generation phase and \textbf{Function Implementation} during the execution phase, which is shown in Figure \ref{fig:framework}.
During the function declaration stage, \ours defines the function name and formal parameters, utilizing them as prompts to generate code. 
Subsequently, in the function implementation stage, \ours implements the declared functions, serving for the direct execution of the generated code.
By defining different functions, \ours can support reasoning over data from various modalities, making it a flexible and scalable framework.
Importantly, \ours eliminates the need to convert different modalities of data into a single modality beforehand. 
Instead, it acquires information within the origin modal by the functions themselves. 
To the best of our knowledge, \ours is the first work to explore the power of LLMs in handling heterogeneous data without requiring domain-specific retrieval or modal transformation.
Experiments demonstrate that \ours significantly outperforms previous
methods.

In summary, our contributions are as follows:
\begin{itemize}
    \item We introduce \ours, a program-based prompting framework that enables reasoning over heterogeneous data without domain-specific retrieval and modal transformation.
    \item We implement a few-shot code generation and execution pipeline, calling various functions by function declaration and implementation to perform information-seeking across data from different sources and modalities.
    \item Experiments show the effectiveness that \ours achieves the best performances under the few-shot settings on HybridQA \cite{chen2020hybridqa} and achieves state-of-the-art performances under all settings on MultiModalQA \cite{talmor2020multimodalqa}. 
\end{itemize}

\section{Method}
\subsection{\ours Framework}
\label{sec:hpropro_framework}

\paragraph{Task Formulation}
In this paper, our focus is on the task of hybrid question answering, which involves answering questions based on heterogeneous information sources such as tables, text, and images. 
The objective is to provide accurate answers to questions based on the given heterogeneous data.
Figure \ref{fig:framework} provides the comparison between retrieval-based methods and our proposed approach \ours. 
Similar to existing program-based prompting approaches, \ours follows a paradigm that involves generating code and executing it to obtain the final answer. 
Unlike the previous approaches with a separate retriever, we deal with the input data with external functions but not the retriever module.
As a result, we introduce two key components: function declaration and function implementation, which are required during the code generation stage and code execution stage, respectively.
In the following sections, we will delve into both parts of the framework and discuss their roles and functionalities.

\paragraph{Function Declaration}

The function declaration process in \ours serves the purpose of defining appropriate functions that can be utilized during the code generation phase. During this stage, it is necessary to specify the function name and formal parameters. These declared functions are treated as input prompts for LLMs and are expected to be leveraged to generate code.
In Figure \ref{fig:framework}, the functions with different highlight backgrounds on the left represent the declared functions. Each function has a specific role, which is described briefly alongside the table and query as the input prompts. These prompts are then fed into LLMs to generate the corresponding code. The LLMs will attempt to utilize the declared functions to generate the desired code.
By providing function declarations as prompts, \ours enables the LLMs to have a better understanding of the expected structure and behavior of the code to be generated. 
This allows for more accurate and controllable code generation, ultimately facilitating the HQA task.

\paragraph{Function Implementation}
The generated code contains formally defined functions, rendering it incapable of direct execution. 
Consequently, the process of function implementation aims to implement the declared functions to make the code able to be executed by off-the-shelf interpreters.
As discussed in Section \ref{sec:intro}, functions are expected to interact with data from various sources. 
However, conventional function structures cannot be accommodated in some scenarios, such as extracting information over unstructured texts or images.
Therefore, we proceed with the initial implementation of the declared functions integrated with the ability of LLMs to ensure that each function encompasses comprehensive functionality.
Specifically, to achieve this process, we pre-design function-related prompts, which are expected to be fed into LLMs when executing the generated code. 

\begin{figure*}[t]
    \centering
    \includegraphics[width=0.8\linewidth]{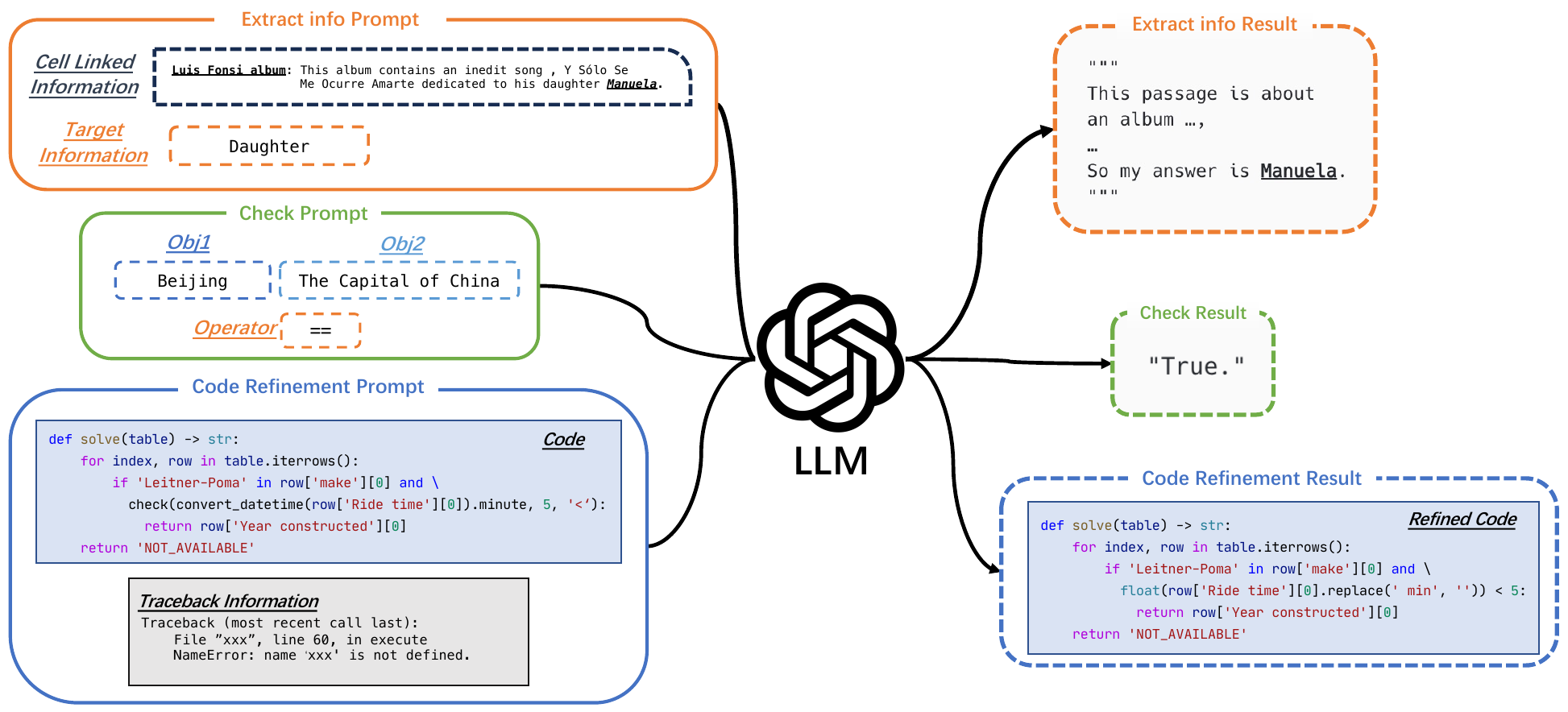}
    \caption{Details of the process of the defined functions and the code refinement.}
    \label{fig:functions}
\end{figure*}

\subsection{Function Instantiation}
\label{sec:function_instantiation}

In this section, we introduce several functions and elaborate on the declaration and implementation of each function to support \ours.

\paragraph{Extract information from external source}
To facilitate reasoning across heterogeneous information sources, we introduce the function "\texttt{extract\_info}". 
Since data from these sources is often unstructured, the process of extracting information can be likened to a reading comprehension task.
In the function declaration, "\texttt{extract\_info}" is defined as "\texttt{extract\_info(cell, target\_information)}". 
Here, "\texttt{cell}" refers to a specific cell in a table, and "\texttt{target\_information}" represents the information that is required to be extracted, as shown in Figure \ref{fig:functions}. 
The function's purpose is to extract the relevant information from the paragraph or image associated with the "\texttt{cell}" based on the specified "\texttt{target\_information}". It should return the extracted information as a textual string.
All the necessary information, including the function name and its parameters, will be part of the generated code and are expected to be generated by LLMs. 
During the function implementation process, we utilize an automatically constructed dictionary to locate the corresponding paragraphs or images based on the provided cell. Subsequently, we construct prompts to invoke LLMs based on the data types, which can be categorized into text-based extraction and image-based extraction.
The detailed prompts are introduced in Appendix \ref{sec:appendix-all-prompts}$.$

\paragraph{Compare two pieces of information}

Code often includes rich comparisons between two objects using operators like "\texttt{>}", "\texttt{<}", or "\texttt{==}". 
However, when dealing with heterogeneous data, the information extracted from various sources may not adhere to a strict format. 
The form of information obtained from functions like "\texttt{extract\_info}" cannot be predetermined. 
As a result, the traditional comparison operators cannot be directly applied to compare two objects, such as comparing the values $"20,000"$ and "ten thousand", or comparing "Beijing" and "the capital of China".
To address this issue, we propose a more flexible function called "\texttt{check}". 
In the function declaration process, "\texttt{check}" can be defined as "\texttt{check(obj1, obj2, op)}". 
As shown in Figure \ref{fig:functions}, "\texttt{obj1}" and "\texttt{obj2}" are two strings representing pieces of information. 
These strings can be the contents of table cells, information obtained from other functions, or directly generated by LLMs based on natural language questions. 
The "\texttt{op}" parameter represents one of three operators: "\texttt{>}", "\texttt{<}", or "\texttt{==}".
The purpose of the "\texttt{check}" function is to compare whether "\texttt{obj1}" and "\texttt{obj2}" are semantically consistent under the specified "\texttt{op}" operator. 
In other words, it evaluates if the semantic relationship between the two objects aligns with the given operator.
Similar to the "\texttt{extract\_info}" function, all the relevant information, including the function name and its actual parameters, will be part of the generated code and are expected to be generated by LLMs. 
During the function implementation process, we provide some few-shot cases as prompts to guide LLMs on how to compare the objects.
The detailed prompts are introduced in Appendix \ref{sec:appendix-all-prompts}$.$

\subsection{Code Refinement}
\label{sec:code_refinement}
In \ours, the final answer is obtained by executing the generated code using a standard Python interpreter. 
Any error in the code will terminate the execution process.
However, since the model cannot predict the results returned by each function during code generation, there is a possibility that the model may generate code with mismatched processing methods. 
This can lead to execution errors or empty results when running the code.
Since initial outputs from LLMs can be improved through iterative feedback and refinement \cite{madaan2023self}, we perform code refinement by re-calling the LLMs and incorporating error codes and traceback information into the prompts to generate new code.
Figure \ref{fig:functions} illustrates the prompts used for code refinement. By providing the above information to LLMs, the models are expected to reconsider the code generation process and generate new code that can alleviate the issues encountered.
The detailed prompts are introduced in Appendix \ref{sec:appendix-all-prompts}$.$

\begin{figure*}[t]
    \centering
    \includegraphics[width=1.0\linewidth]{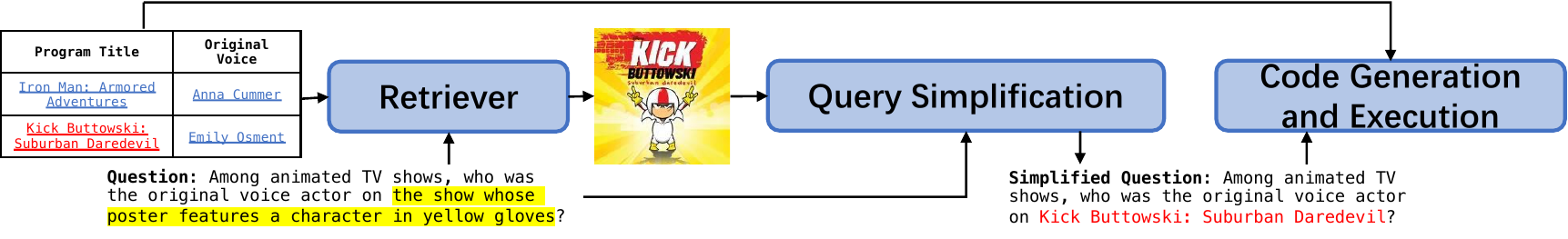}
    \caption{Schematic diagram of query simplification process.}
    \label{fig:query_simplify}
\end{figure*}

\subsection{Query Simplification}

In the HQA task, code generation is often performed based on the input of a table and a question since including all relevant data as input would result in an extensive input length.
However, the reasoning process often involves linked passages or images, which are not directly visible during the code generation phase. This increases the burden of the code generation process.
To address this issue, we employ query simplification to simplify the question and establish links between the question and the table cells before conducting code generation. 
Figure \ref{fig:query_simplify} illustrates the schematic diagram of the query simplification process.
Taking \emph{"Among animated TV shows, who was the original voice actor on the show whose poster features a character in yellow gloves"} as an example, we utilize a general retriever\footnote{The general retriever stands for either a naive retriever or a neural retriever trained on a general corpus, rather than a customized retriever trained on a specific task.} initially to retrieve relevant information from passages or images in the hyperlinks. 
Query simplification involves using LLMs that take as input the retrieved passage or image, the original question, and the table. 
The goal is to replace the span in the question (such as \emph{"the show whose poster features a character in yellow gloves"}) with the corresponding content in the table cell (such as \emph{"Kick Buttowski: Suburban Daredevil"}). 
The detailed prompts are introduced in Appendix \ref{sec:appendix-all-prompts}$.$

\section{Experiments}
\subsection{Datasets}
We conduct experiments on two typical HQA datasets: HybridQA\cite{chen2020hybridqa} and MultiModalQA\cite{talmor2020multimodalqa}. Both datasets involve the task of mixed reasoning over diverse sources of data.
HybridQA necessitates reasoning over hybrid contexts that consist of both tables and texts. On the other hand, MultiModalQA requires reasoning over tables, texts, and images.
To evaluate \ours, we follow the official evaluation metrics provided by the datasets. We report the exact match and F1 score on both HybridQA and MultiModalQA.
For more detailed statistics about the datasets, please refer to Appendix \ref{sec:appendix_statistics}.

\subsection{Experimental Settings}
In all our experiments, we utilize different versions of the GPT language models for different components. 
Specifically, we use \texttt{gpt-4-0613} as the backbone model for code generation, code refinement, and query simplification.
For implementing the function "\texttt{extract\_info}", we employ \texttt{gpt-4-0613} and \texttt{gpt-4-vision-preview} to extract information in the passages and images respectively.
For implementing the function "\texttt{check}", we employ \texttt{gpt-3.5-turbo}.
In the process of query simplification, for the HybridQA task, we employ a hybrid retriever that combines TF-IDF and longest-substring matching \cite{chen2020hybridqa} as the retriever.
For the MultiModalQA task, we utilize sentence transformers \cite{reimers-2020-multilingual-sentence-bert} as the retriever respectively.
The temperature parameter for all models is set to $0$.
All few-shot experiments for code generation are in the settings of $4$ shots.
Furthermore, in the oracle settings of MultiModalQA, where the golden passage and image are provided, we remove the query simplification module. This allows us to directly feed all relevant data to the language models without encountering issues related to excessive input length.

\subsection{Baseline Systems}
We compare \ours to various methods on HybridQA and MultiModalQA, which can be mainly divided into with(\textit{w.}) and without(\textit{w.o.}) fine-tuning approaches. 
For HybridQA, approaches \textit{w.} fine-tuning stand for the method that trained on the training set, including MAFiD \cite{lee2023mafid}, S$^3$HQA \cite{lei2023s}, etc, and approaches \textit{w.o.} fine-tuning include the Unsupervised-QG \cite{pan2021unsupervised} and End-to-End QA with retriever on GPT-4.
For MultiModalQA, baseline methods consist of approaches \textit{w.} fine-tuning including SKURG\cite{yang2023enhancing}, PReasM-Large\cite{yoran2022turning}, etc,
and approaches \textit{w.o.} fine-tuning including Binder\cite{cheng2022binding}, MMHQA-ICL \cite{liu2023mmhqa}, etc.

\begin{table*}[t]
    \centering
    \small
    \scalebox{0.86}{
    \begin{tabular}{p{4.4cm}cccccccccccc}
    \toprule
        \multirow{4}{*}{\textbf{Models}}
        & \multicolumn{4}{c}{\textbf{Table}} & \multicolumn{4}{c}{\textbf{Passage}} & \multicolumn{4}{c}{\textbf{Total}} \\
        \cmidrule(lr){2-5}
        \cmidrule(lr){6-9}
        \cmidrule(lr){10-13}
        & \multicolumn{2}{c}{Dev} & \multicolumn{2}{c}{Test} & \multicolumn{2}{c}{Dev} & \multicolumn{2}{c}{Test} & \multicolumn{2}{c}{Dev} & \multicolumn{2}{c}{Test} \\
        \cmidrule(lr){2-3}
        \cmidrule(lr){4-5}
        \cmidrule(lr){6-7}
        \cmidrule(lr){8-9}
        \cmidrule(lr){10-11}
        \cmidrule(lr){12-13}
        & EM & F1 & EM & F1 & EM & F1 & EM & F1 & EM & F1 & EM & F1 \\
        \midrule
        \multicolumn{13}{c}{\textit{Approaches w. Fine-tuning }} \\
        HYBRIDER \cite{chen2020hybridqa} & $54.3$ & $61.4$ & $56.2$ & $63.3$ & $39.1$ & $45.7$ & $37.5$ & $44.4$ & $44.0$ & $50.7$ & $43.8$ & $50.6$ \\ 
        DocHopper \cite{sun2021end} & -- & -- & -- & -- & -- & -- & -- & -- & $47.7$ & $55.0$ & $46.3$ & $53.3$ \\
        MuGER$^2$ \cite{wang2022muger2} & $60.9$ & $69.2$ & $58.7$ & $66.6$ & $56.9$ & $68.9$ & $57.1$ & $68.6$ & $57.1$ & $67.3$ & $56.3$ & $66.2$ \\ 
        POINTR \cite{eisenschlos2021mate} & $68.6$ & $74.2$ & $66.9$ & $72.3$ & $62.8$ & $71.9$ & $62.8$ & $71.9$ & $63.4$ & $71.0$ & $62.8$ & $70.2$ \\ 
        DEHG \cite{feng2022multi} & -- & -- & -- & -- & -- & -- & -- & -- & $65.2$ & $\mathbf{76.3}$ & $63.9$ & $\mathbf{75.5}$ \\
        MITQA \cite{kumar2021multi} & $68.1$ & $73.3$ & $68.5$ & $74.4$ & $66.7$ & $75.6$ & $64.3$ & $73.3$ & $65.5$ & $72.7$ & $64.3$ & $71.9$ \\ 
        MAFiD \cite{lee2023mafid} & $69.4$ & $75.2$ & $68.5$ & $74.9$ & $66.5$ & $75.5$ & $65.7$ & $75.3$ & $66.2$ & $74.1$ & $65.4$ & $73.6$ \\
        S$^3$HQA \cite{lei2023s} & $\mathbf{70.3}$ & $\mathbf{75.3}$ & $\mathbf{70.6}$ & $\mathbf{76.3}$ & $\mathbf{69.9}$ & $\mathbf{78.2}$ & $\mathbf{68.7}$ & $\mathbf{77.8}$ & $\mathbf{68.4}$ & $75.3$ & $\mathbf{67.9}$ & $\mathbf{75.5}$ \\
        \midrule
        \multicolumn{13}{c}{\textit{Approaches w.o. Fine-tuning}} \\
        Unsupervised-QG \cite{pan2021unsupervised} & -- & -- & -- & -- & -- & -- & -- & -- & $25.7$ & $30.5$ & -- & -- \\
        GPT-4 End-to-End QA \textit{w.} Retriever & $50.0$\textdagger & $\mathbf{61.8}$\textdagger & -- & -- & $11.1$\textdagger & $13.3$\textdagger & -- & -- & $24.5$\textdagger & $30.0$\textdagger & -- & -- \\
        \ours & $\mathbf{51.4}$\textdagger & $55.9$\textdagger & $\mathbf{52.9}$ & $\mathbf{57.6}$ & $\mathbf{46.8}$\textdagger & $\mathbf{54.4}$\textdagger & $\mathbf{46.5}$ & $\mathbf{57.5}$ & $\mathbf{48.0}$\textdagger & $\mathbf{54.6}$\textdagger & $\mathbf{48.7}$ & $\mathbf{57.7}$ \\
  
        \bottomrule
    \end{tabular}
    }
    \caption{Experimental results on HybridQA. \textdagger \ stands for running on $200$ sampled cases from the validation set. 
    }
    \label{tab:hybridqa_result}
\end{table*}

\subsection{Main Results}
\paragraph{Results on HybridQA}

According to the results presented in Table \ref{tab:hybridqa_result}, it is evident that \ours outperforms all baseline systems among approaches \textit{w.o.} fine-tuning.
GPT-4 End-to-End QA \textit{w.} Retriever stands for leveraging GPT-4 to generate answers directly along with a retriever. To conduct this experiment, we follow the retrieval approach proposed by \citet{chen2020hybridqa}.
In comparison to GPT-4 End-to-End QA \textit{w.} Retriever, \ours achieves more than a $20\%$ improvement in both EM and F1 scores. This result demonstrates the effectiveness of \ours compared with the approaches relied on retrievers.
However, it is important to note that \ours still exhibits a significant performance gap when compared to the state-of-the-art approaches \textit{w.} fine-tuning. We argue that the main reason for this gap is that these methods are fully trained on the HybridQA dataset.
These systems focus on domain-specific training, which includes training a retriever \cite{wang2022muger2, lei2023s}, ranker \cite{kumar2021multi}, or reasoner \cite{eisenschlos2021mate, lee2023mafid}. 
These domain-specific components may lack flexibility and generalization in handling diverse scenarios.

\paragraph{Results on MultiModalQA}
Table \ref{tab:mmqa_result} summarizes the results obtained on the MultiModalQA dataset, where \ours achieves state-of-the-art performances across all experimental settings.
When considering systems \textit{w.o.} fine-tuning, \ours outperforms the previous system MMHQA-ICL by $4.2\%$ and $0.9\%$ in terms of EM and F1 scores, respectively. 
In comparison to the baseline systems Binder and MMHQA-ICL, which utilize modal transformation modules to convert images into texts, \ours employs various functions to directly extract information from different modalities. This approach avoids information loss and is more suitable for real-world scenarios involving heterogeneous data.
It is important to note that the improvements achieved by \ours are non-trivial, considering that MMHQA-ICL leverages domain-specific fine-tuned classifiers and retrievers to obtain the type and relevant passages of each question respectively, which heavily relies on the distribution of the targeted benchmark. In contrast, \ours is performed without any supervised signals from the training set, resulting in a more universal approach.

In the oracle setting which golden passages and images are provided as the input, \ours achieves comparable results to the previous state-of-the-art system MMHQA-ICL in terms of EM.
Demonstrating that regardless of the retriever (only focus on the reasoning part), the results prove that their work highly relies on the retrievers to gain the performances.
Besides, compared to their approach, \ours follows a code generation and execution paradigm, which provides enhanced interpretability and generalizability.

\begin{table}[t]
    \centering
    \small
    \begin{tabular}{p{4.8cm}cc}
    \toprule
         \textbf{Models} & \textbf{EM} & \textbf{F1} \\
         \midrule
        \multicolumn{3}{c}{\textit{ Approaches w. Fine-tuning}} \\
         Implicit-Decomp \cite{talmor2020multimodalqa} & $48.8$ & $55.5$ \\
         AutoRouting \cite{talmor2020multimodalqa} & $42.1$ & $49.5$\\
         SKURG \cite{yang2023enhancing} & $59.4$ & $63.8$  \\
         PReasM-Large \cite{yoran2022turning} & $59.0$ & $65.5$ \\
         \midrule
        \multicolumn{3}{c}{\textit{Approaches w.o. Fine-tuning}} \\
        Binder \cite{cheng2022binding} & $51.0$ & $57.1$ \\
        MMHQA-ICL \cite{liu2023mmhqa} & $54.8$ & $65.8$ \\
        \ours & $\mathbf{59.0}$ & $\mathbf{66.7}$ \\
        \midrule
        \multicolumn{3}{c}{\textit{Approaches in Oracle Setting}} \\
        Binder$_{oracle}$ \cite{cheng2022binding} & $58.1$ & $64.5$ \\
        MMHQA-ICL$_{oracle}$ \cite{liu2023mmhqa} & $65.0$ & $\mathbf{75.9}$ \\
        \ours & $\mathbf{65.1}$ & $73.1$ \\
        \bottomrule
         
    \end{tabular}
    \caption{Experimental results on MultiModalQA.}
    \label{tab:mmqa_result}
\end{table}

\begin{table}[t]
    \centering
    \small
    \scalebox{0.93}{
    \begin{tabular}{p{3.3cm}cccc}
    \toprule
        \multirow{3}{*}{\textbf{Models}}
        & \multicolumn{2}{c}{\textbf{HybridQA}} & \multicolumn{2}{c}{\textbf{MultiModalQA}} \\
        \cmidrule(lr){2-3}
        \cmidrule(lr){4-5}
    
          & \textbf{EM} & \textbf{F1} & \textbf{EM} & \textbf{F1} \\
         \midrule
         \ours & $\mathbf{48.0}$ & $\mathbf{54.6}$ & $\mathbf{56.0}$ & $\mathbf{62.8}$ \\
         \ \ --\ "\texttt{check}" & $44.5$ & $52.5$ & $35.5$ & $38.0$ \\
         \ \ --\ Question Simplification & $43.0$ & $50.0$ & $37.0$ & $39.4$ \\
         \ \ --\ Code Reflection & $43.5$ & $50.9$ & $54.0$ & $60.5$ \\
        \bottomrule
         
    \end{tabular}
    }
    \caption{Ablation studies on HybridQA and MultiModalQA. All ablation studies are performed on $200$ randomly sampled subsets from validation sets.}
    \label{tab:ablation_studies}
\end{table}

\subsection{Ablation Study}
\paragraph{Effect of the function "\texttt{check}"}
The function "\texttt{check}" is designed to compare the semantic relations between two objects, offering greater flexibility compared to arithmetic operators such as "\texttt{>}", "\texttt{<}", and "\texttt{==}". To demonstrate the effectiveness of the "\texttt{check}" function, we conduct ablation studies by removing its definition in both the function declaration and function implementation processes.
Table \ref{tab:ablation_studies} presents the results of these ablation studies, highlighting the impact of the "\texttt{check}" function. When the "\texttt{check}" function is removed, there is a noticeable drop of $3.5\%$ and $2.1\%$ points in terms of EM and F1 scores, respectively, in the HybridQA dataset. 
Moreover, the removal of the "\texttt{check}" function has an even more substantial impact on the MultiModalQA dataset. Specifically, the results drop by more than $20\%$ for both EM and F1 scores.
This is because constraints from images are weaker than those from passages since LLMs can copy spans from the passages as the answer, which improves the need for the "\texttt{check}" function in this set of experiments.

\paragraph{Effect of query simplification}
The purpose of query simplification is to alleviate the burden of the code generation process by simplifying the question and establishing links between the question and the table cells. 
In Table \ref{tab:ablation_studies}, we present the effectiveness of query simplification on both the HybridQA and MultiModalQA datasets.
When query simplification is removed, the results demonstrate a decrease of approximately $2\%$ on the HybridQA dataset and a substantial drop of about $20\%$ on the MultiModalQA dataset. 
These findings highlight the effectiveness of query simplification in the HQA task.
It is important to note that the removal of the query simplification module leads to a significant drop specifically in the MultiModalQA dataset. 
We posit that this drop is due to the presence of passages and images that are necessary to answer the question but are not linked in the table, which couldn't be accessed by the model from the prompt. 
Therefore, performing query simplification becomes crucial in handling such scenarios in the MultiModalQA task.

\paragraph{Effect of code refinement}

The code refinement module aims to enable LLMs to reconsider the code generation process based on previous execution traceback. 
In Table \ref{tab:ablation_studies}, we can observe the impact of removing the code refinement module on both the HybridQA and MultiModalQA datasets.
When the code refinement module is removed, there is a noticeable decrease in performance. 
In the HybridQA dataset, the EM and F1 scores drop by $4.5\%$ and $3.7\%$ respectively. 
Similarly, in the MultiModalQA dataset, the EM and F1 scores drop by $2.0\%$ and $2.3\%$ respectively. 
The drop in performance demonstrates the effectiveness of the code refinement module in \ours. By enabling LLMs to refine their code generation process based on previous execution errors, the code refinement module plays a vital role in generating accurate code, thereby enhancing the overall ability of \ours in the HQA task.

\begin{figure}[t]
    \centering
    \includegraphics[width=0.90\linewidth]{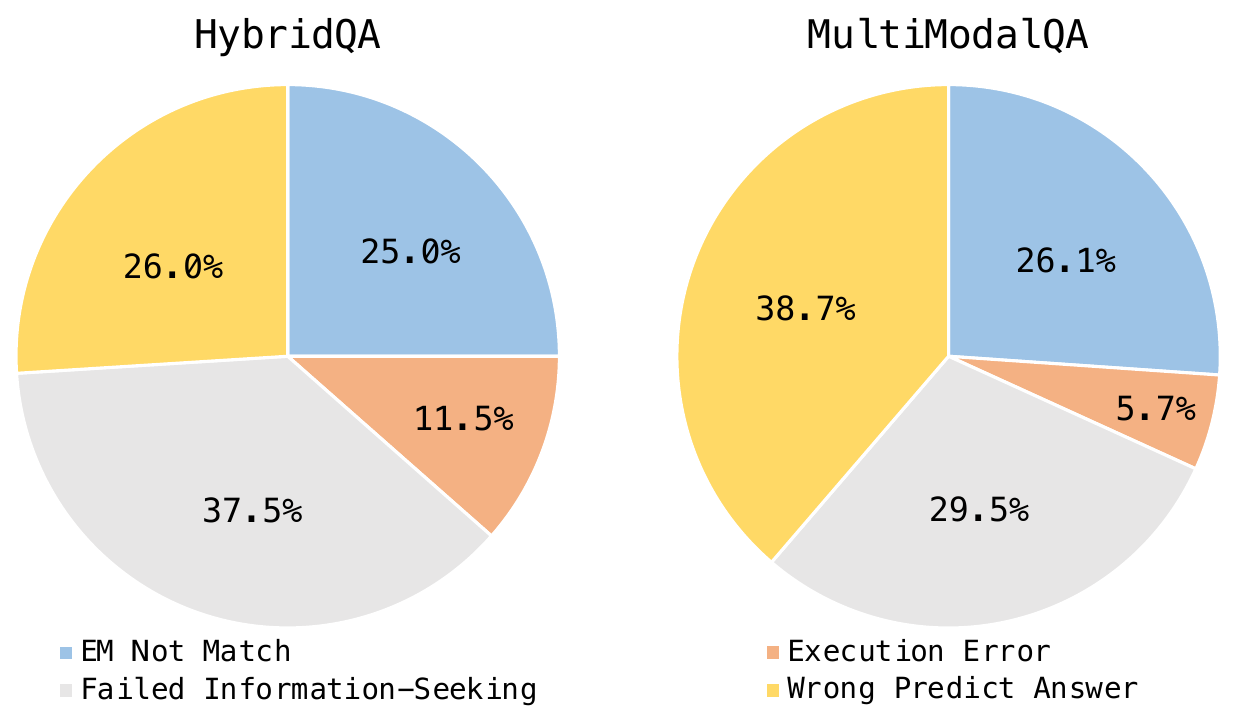}
    \caption{Error percentage of \ours on HybridQA and MultiModalQA.}
    \label{fig:error_percentage}
\end{figure}

\subsection{Error Analysis}
We analyze the errors that occurred within randomly selected subsets of $200$ cases from the validation sets of HybridQA and MultiModalQA.  Our examination reveals that the main errors can be classified into four distinct types, with the corresponding percentages depicted in Figure \ref{fig:error_percentage}.

The first type of error involves predicted answers that possess similar meanings to the golden answers but differ in their expressions ($25.0\%$ for HybridQA and $26.1\%$ for MultiModalQA). 
For instance, an instance may present the predicted answer as "the Southeastern Conference (SEC)", while the correct answer is simply "Southeastern Conference". 
From a technological perspective, we contend that such cases have already been resolved, as the underlying code solution is entirely accurate.

The second type of error observed is related to execution failure ($11.5\%$ for HybridQA and $5.7\%$ for MultiModalQA). 
This error arises due to the inherent complexity of the heterogeneous data, which lacks a standardized format and therefore cannot be effectively addressed using a uniform solution.

The third type of error pertains to failures in the information-seeking from heterogeneous data sources ($37.5\%$ for HybridQA and $29.5\%$ for MultiModalQA). 
These errors occur when the "\texttt{extract\_info}" function fails to produce a valid result. 
This may be attributed to a mismatch between the generated code and the expected solution for answering the given question, or it could be indicative of instability in the implementation of the "\texttt{extract\_info}" function.

The last type of error involves wrong predicted answers ($26.0\%$ for HybridQA and $38.7\%$ for MultiModalQA). 
Due to the similarity between contents in different columns, the model encounters difficulty in discerning the appropriate location to locate the answer when generating code solely based on the provided table. 
Addressing this challenge remains a topic for future research.

For the detailed visualization results of this analysis, please refer to Appendix \ref{sec:appendix-error-analysis}.

\section{Related Work}
\subsection{Hybrid Question Answering}
The first line of our related work introduces the HQA task, which focuses on answering questions that require reasoning over diverse information sources. Currently, HQA can be broadly categorized into three subtasks based on the nature of the information sources: table-text question answering \cite{chen2020hybridqa,chen2020open,zhu2021tat}, image-text question answering \cite{reddy2022mumuqa,singh2021mimoqa}, and table-image-text question answering \cite{hannan2020manymodalqa,talmor2020multimodalqa}.
Numerous approaches have been explored for reasoning over heterogeneous data in the context of HQA. Many of these methods primarily focus on supervised fine-tuning over specific benchmarks. This includes training dedicated retrievers \cite{wang2022muger2,kumar2021multi,lei2023s}, rankers \cite{kumar2021multi}, reasoners \cite{wang2022muger2,kumar2021multi,eisenschlos2021mate,lee2023mafid,lei2023s}, or transforming different modalities of data into a unified modality \cite{cheng2022binding,liu2023mmhqa,li2021tsqa}.
In contrast to existing works, \ours performs reasoning over heterogeneous data without relying on domain-specific retriever and modal transformation modules. Instead, it integrates various functions to facilitate information-seeking across data from different sources and modalities.

\subsection{Program-based Prompting}

The second line of our related work focuses on the program-based prompting strategy, with two closely related works: Program-of-Thought-Prompting \cite{chen2022program,gao2023pal} and Binder \cite{cheng2022binding}.
Program-of-Thought-Prompting \cite{chen2022program,gao2023pal} generates code and executes it using an interpreter. However, their approach is not designed to handle heterogeneous data. In contrast, \ours integrates function declaration and implementation to specify different functions, enabling effective handling of heterogeneous data.
On the other hand, Binder \cite{cheng2022binding} converts images into passages and pre-retrieves relevant passages. These passages, along with the table and question, are then fed into LLMs to generate SQL and Python code for solving the question. In comparison, \ours does not rely on a modal transformation module or a retriever. Instead, it utilizes various functions to directly interact with data from different sources and modalities.

\section{Conclusion}
In this work, we propose \ours, a novel program-based prompting framework for HQA tasks, which does not require domain-specific retriever and modal transformation, but integrates various functions to interact with heterogeneous data instead. 
Experimental results on two typical HQA benchmarks HybridQA and MultiModalQA show the effectiveness of \ours that \ours achieves the best performances under the few-shot settings. 
For future work, we hope to further utilize the coding capabilities of the LLMs, allowing the model to judge and self-create more customized functions based on different scenarios.

\section*{Limitations}
The main limitation of this paper is that the performance of \ours relies on the abilities of LLMs, which vary according to the different choices of LLMs. 
Model updates and server status may affect our experimental results.
In addition, the existing benchmarks only focus on heterogeneous data containing tables, passages, and images. More types of data including knowledge graphs and charts are expected to be explored in the future.

\section*{Ethics Statement}
In this paper, we propose \ours, a program-based prompting framework for the HQA task. We conduct experiments on two benchmarks, namely, HybridQA and MultiModalQA. Both datasets are free and open for research use, which means no ethics issues. 

\bibliography{custom}

\appendix

\section{Detail Prompts of of \ours}
\label{sec:appendix-all-prompts}
\begin{figure}[h]
    \centering
    \includegraphics[width=1.0\linewidth]{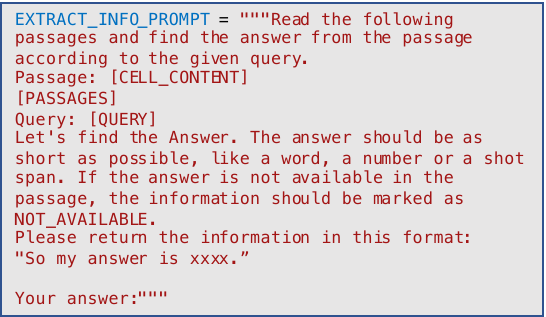}
    \caption{Details of the prompt of "\texttt{extract\_info}" function.}
    \label{fig:appendix_extract_info_prompt}
\end{figure}

\begin{figure}[h]
    \centering
    \includegraphics[width=1.0\linewidth]{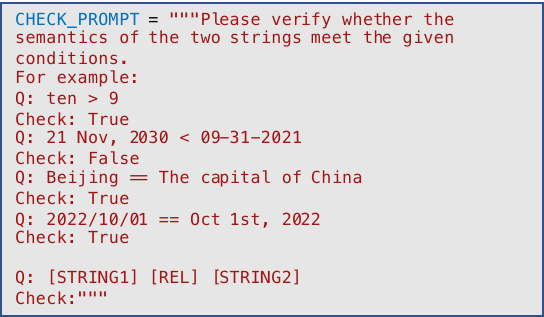}
    \caption{Details of the prompt of "\texttt{check}" function.}
    \label{fig:appendix_check_prompt}
\end{figure}

\begin{figure}[h]
    \centering
    \includegraphics[width=1.0\linewidth]{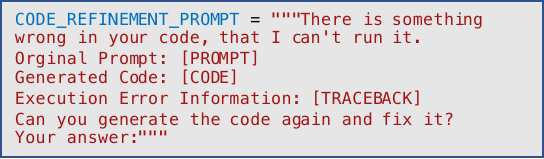}
    \caption{Details of the prompt of the code refinement process.}
    \label{fig:appendix_code_refinement_prompt}
\end{figure}

\begin{figure}[h]
    \centering
    \includegraphics[width=1.0\linewidth]{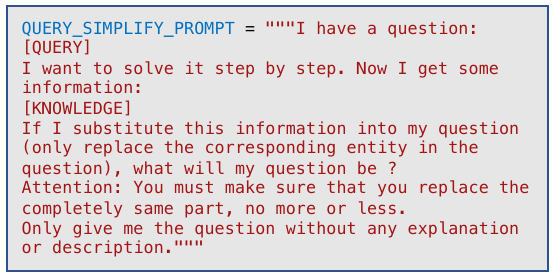}
    \caption{Details of the prompt of the query simplification process.}
    \label{fig:appendix_query_simplify_prompt}
\end{figure}

The system prompt and detail prompts of the function "\texttt{extract\_info}", function "\texttt{check}", code refinement, and query simplification are shown in Figure \ref{fig:appendix_extract_info_prompt}, Figure \ref{fig:appendix_check_prompt},  Figure \ref{fig:appendix_code_refinement_prompt} and Figure \ref{fig:appendix_query_simplify_prompt}.

\section{Data Statistics for Each Dataset}
\label{sec:appendix_statistics}
The statistics of HybridQA and MultiModalQA are presented in Table \ref{tab:appendix_hybridqa} and Table \ref{tab:appendix_mmqa}.

\section{Error Analysis}
\label{sec:appendix-error-analysis}
The error analysis results are presented in Figure \ref{tab:appendix-error-analysis}.

\begin{figure}[h]
    \centering
    \includegraphics[width=1.0\linewidth]{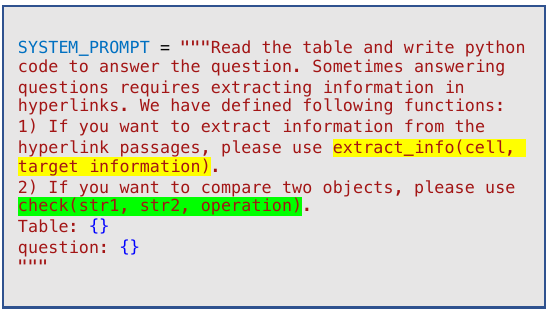}
    \caption{Details of the system prompt.}
    \label{fig:appendix_system_prompt}
\end{figure}

\begin{table}[h]
\small
\centering
\begin{tabular}{lccccc}
\toprule
Split & Train & Dev & Test & Total \\
\midrule
In-Passage  &  35,215 & 2,025  & 20,45  & 39,285 (56.4\%) \\
In-Table & 26,803 & 1,349 & 1,346 & 29,498 (42.3\%)  \\
Computed & 664  & 92 & 72 & 828 (1.1\%) \\
Total & 62,682 & 3,466 & 3,463 & 69,611 \\
\bottomrule
\end{tabular}
\caption{Data Statistics for HybridQA. }
\label{tab:appendix_hybridqa}
\vspace{-2ex}
\end{table}

\begin{table}[h]\small
\centering
\resizebox{0.5 \textwidth}{!}{
\begin{tabular}{ll} 
\hline
{\emph{Measurement}} & {\emph{Value}} \\ 
\hline 
\# Distinct Questions & 
29,918 
\\
Train multimodal questions & 
34.6\% 
\\
Dev.+test multimodal questions & 
40.1\%
\\
Train compositional questions & 
58.8\% 
\\
Dev.+test compositional questions & 
62.3\% 
\\
Average question length~(words) & 
18.2 
\\
Average \# of answers per question & 
1.16 
\\
List answers &
7.4\%  
\\
List answers per intermediate question & 
18.9\%  
\\
Average answer length (words) & 
2.1 
\\
\# of distinct words in questions & 
49,649 
\\
\# of distinct words in answers & 
20,820 
\\ 
\# of distinct context tables & 
11,022 
\\ 
\hline
\end{tabular}}
\caption{Data statistics for MultiModalQA.}
\label{tab:appendix_mmqa}

\end{table}

\begin{table*}[h]
    \centering
    \small
    \scalebox{0.85}{
    \begin{tabular}{m{2.1cm}m{3.0cm}m{6.3cm}m{5.8cm}}
        \toprule
        \textbf{Type} & \textbf{Question} & \textbf{Generated Code} & \textbf{Result Comparison} \\

        \midrule

        EM Not Match & 
        What conference does the team won in 2016 belong to? & 
        \includegraphics[width=6.2cm]{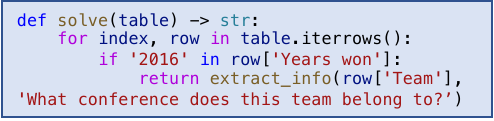}\xspace &
       The predicted answer is "the Southeastern Conference ( SEC )", while the golden answer is "Southeastern Conference". The code has the completely correct solution but has different expressions with the golden answer. \\ 

        \midrule

        Execution Error & Who was the older person involved in writing the book from 2000? & \includegraphics[width=6.2cm]{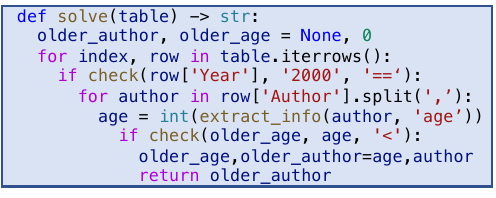} & The golden answer is "Sally Jenkins", while the existence of function \texttt{"split"} and \texttt{"int"} make the code fail to execute since the heterogeneous data is complex. \\

        \midrule

        Failed Information Seeking & What shipbuilder designed the ship that the delivery voyage was the longest by a patrol boat? & \includegraphics[width=6.2cm]{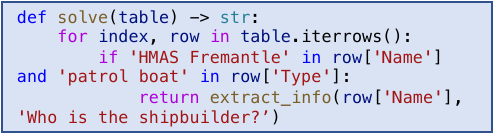}\xspace & The golden answer is "Brooke Marine", while the code cannot locate the answer since the "\texttt{extract\_info}" returns "None". \\
        
        \midrule
        
        Wrong Predicted Answer & At which air force base did the pilot who graduated in 1968 serve ? & \raisebox{-5pt}{\includegraphics[width=6.2cm]{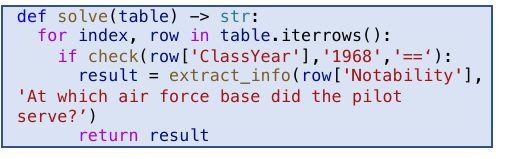}}\xspace & The prediction answer is "Hurlburt Field", while the golden answer is "Hickam". The model cannot distinguish where to find the answer between the columns "ClassYear" and "Notability" since the information in the two columns is similar. \\

        \bottomrule
    \end{tabular}
    }
    \caption{Main error types of \ours in HybridQA and MultiModalQA. }
    \label{tab:appendix-error-analysis}
\end{table*}

\end{document}